\pdfoutput=1
\documentclass[10pt,twocolumn,letterpaper]{article}

\usepackage{cvpr}

\usepackage{times}
\usepackage{epsfig}
\usepackage{url}
\usepackage{graphicx}
\usepackage{amssymb}
\usepackage{amsmath}
\usepackage{tabularx}
\usepackage{multirow}
\usepackage{booktabs}
\usepackage{mathtools}
\usepackage{enumerate}

\usepackage{color}

\usepackage[table]{xcolor}

\def\eg{\textit{e.g.,~}}
\def\cf{\textit{cf.~}}
\def\ie{\textit{i.e.~}}
\def\wrt{\textit{w.r.t.~}}
\def\etal{\textit{et~al.~}}

\usepackage{subfigure}
\usepackage[font=small,labelfont=bf,tableposition=top]{caption}
\definecolor{lgray}{RGB}{225,225,225}
\definecolor{mgray}{RGB}{185,185,185}
\definecolor{lightgray}{gray}{0.40}

\usepackage[pagebackref=true,breaklinks=true,letterpaper=true,colorlinks,bookmarks=false]{hyperref}

\cvprfinalcopy

\def\cvprPaperID{2471}

\ifcvprfinal\fi
\begin{document}

\title{\Large \bf Virtual Worlds as Proxy for Multi-Object Tracking Analysis}

\newcommand\Mark[1]{\textsuperscript#1}

\author{ 
\vspace*{-2mm}
Adrien Gaidon\Mark{1}\thanks{AG and QW have contributed equally}
\qquad
Qiao Wang\Mark{2}\footnotemark[1]
\qquad
Yohann Cabon\Mark{1}
\qquad
Eleonora Vig\Mark{1}\thanks{EV is currently at the German Aerospace Center}\\
\\
\Mark{1}{\small Computer Vision group, Xerox Research Center Europe,
France}\\
\Mark{2}{\small School of Electrical, Computer, and Energy Engineering and School
of Arts, Media, and Engineering, Arizona
State University, USA}\\
{\tt\small \{adrien.gaidon,yohann.cabon\}@xrce.xerox.com \ 
qiao.wang@asu.edu \ eleonora.vig@dlr.de}\\
{\tt\small
\url{http://www.xrce.xerox.com/Research-Development/Computer-Vision/Proxy-Virtual-Worlds}}
}

\maketitle

\vspace*{-7mm}
\begin{abstract}
\vspace*{-2mm}
Modern computer vision algorithms typically require expensive data acquisition
and accurate manual labeling.
In this work, we instead leverage the recent progress in computer graphics to
generate fully labeled, dynamic, and photo-realistic proxy virtual worlds.
We propose an efficient real-to-virtual world cloning method, and validate our
approach by building and publicly releasing a new video dataset, called
``Virtual
KITTI''~\footnote{\label{vkitti-url}\url{http://www.xrce.xerox.com/Research-Development/Computer-Vision/Proxy-Virtual-Worlds}},
automatically labeled with accurate ground truth for object detection,
tracking, scene and instance segmentation, depth, and optical flow.
We provide quantitative experimental evidence suggesting that (i) modern deep
learning algorithms pre-trained on real data behave similarly in real and
virtual worlds, and (ii) pre-training on virtual data improves performance.
As the gap between real and virtual worlds is small, virtual worlds enable 
measuring the impact of various weather and imaging conditions on recognition
performance, all other things being equal.
We show these factors may affect drastically otherwise high-performing deep
models for tracking.
\vspace*{-5mm}
\end{abstract}

\section{Introduction}

\begin{figure}
\vspace*{-2.5mm}
\center
\includegraphics[width=0.47\textwidth]{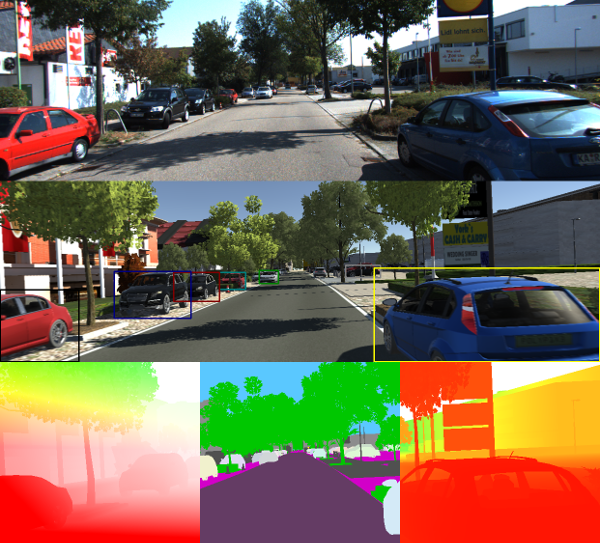} \
\vspace*{-6mm}
\caption{\label{fig:frame_only_virtual_vs_real} Top: a frame of a video
from the KITTI multi-object tracking benchmark~\cite{Geiger2012}. Middle: the
corresponding rendered frame of the synthetic clone from our Virtual KITTI
dataset with automatic tracking ground truth bounding boxes. Bottom:
automatically generated ground truth for optical flow (left), scene- and
instance-level segmentation (middle), and depth (right).} 
\vspace*{-4.5mm} 
\end{figure}

Although cheap or even no annotations might be used at training time via
weakly-supervised (resp. unsupervised) learning, experimentally evaluating the
generalization performance and robustness of a visual recognition model
requires accurate full labeling of large representative datasets.
This is, however, challenging in practice for video understanding tasks like
multi-object tracking (MOT), because of the high data acquisition and labeling
costs that limit the quantity and variety of existing video benchmarks.
For instance, the KITTI~\cite{Geiger2012} multi-object tracking benchmark contains
only 29 test sequences captured in similar good conditions and from a single
source.
To the best of our knowledge, none of the existing benchmarks in computer
vision contain the minimum variety required to properly assess the performance
of video analysis algorithms: varying conditions (day, night, sun, rain, \ldots),
multiple detailed object class annotations (persons, cars, license plates, \ldots),
and different camera settings, among many others factors.

Using synthetic data should in theory enable full control of the
data generation pipeline, hence ensuring lower costs, greater
flexibility, and limitless variety and quantity.
In this work, we leverage the recent progress in computer graphics (especially
off-the-shelf tools like game engines) and commodity hardware (especially GPUs)
to \emph{generate photo-realistic virtual worlds used as proxies to assess the
performance of video analysis algorithms}.

Our \emph{first contribution} is a method to generate large, photo-realistic,
varied datasets of synthetic videos, automatically and densely labeled for
various video understanding tasks.
Our main novel idea consists in creating virtual worlds not from scratch, but
by \emph{cloning} a few seed real-world video sequences.
Using this method, our \emph{second and main contribution} is the creation of
the new \textbf{Virtual KITTI dataset} (\cf
Figure~\ref{fig:frame_only_virtual_vs_real}), which at the time of publication
contains 35 photo-realistic synthetic videos (5 cloned from the original
real-world KITTI tracking benchmark~\cite{Geiger2012}, coupled with 7
variations each) for a total of approximately 17,000 high resolution frames,
all with \emph{automatic accurate ground truth for object detection,
tracking, depth, optical flow, as well as scene and instance segmentation at
the pixel level}.

Our \emph{third contribution} consists in quantitatively measuring the
usefulness of these virtual worlds as proxies for multi-object tracking.
We first propose a practical definition of \emph{transferability} of
experimental observations across real and virtual worlds. Our protocol rests on
the comparison of real-world seed sequences with their corresponding synthetic
clones using real-world pre-trained deep models (in particular
Fast-RCNN~\cite{Girshick2015}), hyper-parameter calibration via Bayesian
optimization~\cite{bergstra2013making}, and the analysis of task-specific
performance metrics~\cite{Bernardin2008}.
Second, we validate the usefulness of our virtual worlds for \emph{learning}
deep models by showing that \emph{virtual pre-training followed by real-world
fine-tuning outperforms training only on real world data}.
Our experiments, therefore, suggest that the recent progress in computer graphics
technology allows one to easily build virtual worlds that are indeed
effective proxies of the real world from a computer vision perspective.

Our \emph{fourth contribution} builds upon this small virtual-to-real gap to
measure the potential impact on recognition performance of varied weather
conditions (like fog), lighting conditions, and camera angles, \emph{all other
things being equal}, something impractical or even impossible in real-world
conditions.
Our experiments show that \emph{these variations may significantly deteriorate
the performance of normally high-performing models trained on large real-world
datasets}. This lack of generalization highlights the importance of open
research problems like unsupervised domain adaptation and building more
varied training sets, to move further towards applying computer vision in
the wild.

The paper is organized as follows.
Section~\ref{s:relwork} reviews related works on using synthetic data
for computer vision.
Section~\ref{s:proxy} describes our approach to build virtual worlds in general
and Virtual KITTI in particular.
Section~\ref{s:experiments} reports our multi-object tracking experiments using
strong deep learning baselines (Section~\ref{s:motbaselines}) to assess the
transferability of observations across the real-to-virtual gap
(Section~\ref{s:mottransfer}), the benefits of virtual pre-training
(Section~\ref{s:virtualpretraining}), and the impact of various weather and
imaging conditions on recognition performance (Section~\ref{s:motvarexps}).
We conclude in section~\ref{s:conclusion}.

\section{Related Work}
\label{s:relwork}

Several works investigate the use of 3D synthetic data to tackle standard 2D
computer vision problems such as object detection~\cite{pepik2012teaching},
face recognition, scene understanding~\cite{satkin2012data}, and optical flow
estimation~\cite{sintel}. From early on, computer vision researchers leveraged
3D computer simulations to model articulated objects including human
shape~\cite{grauman2003inferring}, face, and hand
appearance~\cite{ballan2012motion}, or even for scene interpretation and vision
as inverse graphics~\cite{Battaglia2013,Mansinghka2013,Kulkarni2015}.
However, these methods typically require controlled virtual environments, are
tuned to constrained settings, and require the development of task-specific
graphics tools.
In addition, the lack of photorealism creates a significant domain gap
between synthetic and real world images, which in turn might render synthetic
data too simplistic to tune or analyze vision algorithms~\cite{Vaudrey2008}.

The degree of photorealism allowed by the recent progress in computer graphics
and modern high-level generic graphics platforms
enables a more widespread use of synthetic data generated under less
constrained settings.
First attempts to use synthetic data for training are mainly
limited to using rough synthetic models or synthesized real examples (\eg
of pedestrians~\cite{Broggi05,Stark10}).  
In contrast, Mar\'in \etal \cite{Marin10,Vazquez:12,Vazquez2014} went
further and positively answer the intriguing question whether one can
learn appearance models of pedestrians in a virtual world and use the
learned models for detection in the real world.
A related approach is described in~\cite{Hattori15}, but for scene- and
scene-location specific detectors with fixed calibrated surveillance cameras
and a priori known scene geometry.
In the context of video surveillance too, \cite{taylor2007ovvv} proposes a
virtual simulation test bed for system design and evaluation.
Several other works use 3D CAD models for more general object pose
estimation~\cite{Aubry2014,Busto2015} and detection~\cite{Sun2014,Peng2015}.

Only few works use photo-realistic imagery for \emph{evaluation purposes}, and
in most cases these works focus on low-level image and video processing tasks.
Kaneva~\etal~\cite{Kaneva11} evaluate low-level image features, while
Butler~\etal~\cite{Butler12} propose a synthetic benchmark for optical flow
estimation: the popular MPI Sintel Flow Dataset. The recent work of
Chen~\etal~\cite{Chen15} is another example for basic building blocks of
autonomous driving.
These approaches view photo-realistic imagery as a way of obtaining ground
truth that cannot be easily obtained otherwise (\eg optical flow).
When ground-truth can be collected, for instance via crowd-sourcing, real-world
imagery is often preferred over synthetic data because of the artifacts the
latter might introduce.

In this paper, we show that such issues can be partially circumvented using our
approach, in particular for high-level video understanding tasks for which
ground-truth data is tedious to collect. 
We believe current approaches face two major limitations that prevent
broadening the scope of virtual data.  
First, the data generation is itself costly and time-consuming, as it often
requires creating animation movies from scratch. This also limits the quantity
of data that can be generated. An alternative consists in recording scenes from
humans playing video games~\cite{Marin10}, but this faces similar time costs,
and further restricts the variety of the generated scenes.
The second limitation lies in the usefulness of synthetic data as a proxy to
assess real-world performance on high-level computer vision tasks, including
object detection and tracking.
It is indeed difficult to evaluate how conclusions obtained from
virtual data could be applied to the real world in general.

Due to these limitations, only few of the previous works have so far exploited
the full potential of virtual worlds: the possibility to generate endless
quantities of varied video sequences on-the-fly. This would be especially
useful in order to assess model performance, which is crucial for real-world
deployment of computer vision applications.
In this paper, we propose steps towards achieving this goal by
addressing two main challenges: (i) automatic generation of arbitrary
photo-realistic video sequences with ground-truth by scripting modern game
engines, and (ii) assessing the degree of transferability of experimental
conclusions from synthetic data to the real world.

\section{Generating Proxy Virtual Worlds}
\label{s:proxy}

Our approach consists in five steps detailed in the following sections: (i)
the acquisition of a small amount of real-world data as a starting point for
calibration (Section~\ref{s:realdata}), (ii) the ``cloning'' of this real-world
data into a virtual world (Section~\ref{s:synthclone}), (iii) the automatic
generation of modified synthetic sequences with different weather or imaging
conditions (Section~\ref{s:synthmods}), (iv) the automatic generation of
detailed ground truth annotations (Section~\ref{s:synthgt}), and (v) the
quantitative evaluation of the ``usefulness'' of the synthetic data
(Section~\ref{s:transfer}).
We describe both the method and the particular choices made to generate our
Virtual KITTI dataset.

\subsection{Acquiring real-world (sensor) data}
\label{s:realdata}

The first step of our approach consists in the acquisition of a limited amount
of seed data from the real world for the purpose of calibration. Two types of
data need to be collected: videos of real-world scenes and physical
measurements of important objects in the scene including the camera itself.
The quantity of data required by our approach is much smaller than what is
typically needed for training or validating current computer vision models, as
we do not require a reasonable coverage of all possible scenarios of interest.
Instead, we use a small fixed set of core real-world video sequences to
initialize our virtual worlds, which in turn allows one to generate many
varied synthetic videos.
Furthermore, this initial seed real-world data results in higher quality
virtual worlds (\ie closer to real-world conditions) and to quantify their
usefulness to derive conclusions that are likely to transfer to real-world
settings.
 
In our experiments, we use the KITTI dataset~\cite{Geiger2012} to initialize
our virtual worlds. This standard public benchmark was captured from a car
driving in the German city of Karlsruhe, mostly under sunny conditions. The
sensors used to capture data include gray-scale and color cameras, a 3D laser
scanner, and an inertial and GPS navigation system.
From the point clouds captured by the 3D laser scanner, human annotators
labeled 3D and 2D bounding boxes of several types of objects including cars and
pedestrians.
In our experiments we only consider cars as objects of interest for simplicity
and because they are the main category of KITTI.
The annotation data include the positions and sizes of cars, and their rotation
angles about the vertical axis (yaw rotation). The movement of the camera
itself was recorded via GPS (latitude, longitude, altitude) and its orientation
(roll, pitch, yaw) via a GPS/IMU sensor, which has a fixed spatial relationship
with the cameras.

\subsection{Generating synthetic clones}
\label{s:synthclone}

\begin{figure}
\center
\hspace*{-4mm}
\includegraphics[width=0.51\textwidth]{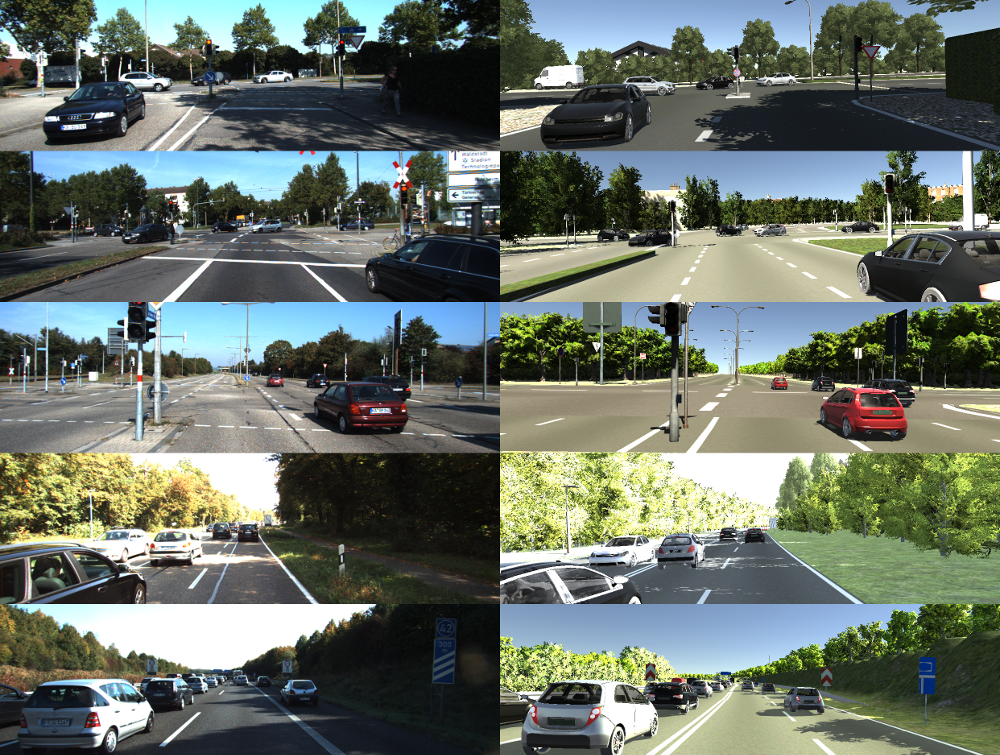} \
\vspace*{-3mm}
\caption{\label{fig:allclones} Frames from 5 real KITTI videos (left, sequences
1, 2, 6, 18, 20 from top to bottom) and rendered virtual clones (right).
} 
\vspace*{-1mm} 
\end{figure}

The next step of our approach consists in \emph{semi-automatically} creating
photo-realistic dynamic 3D virtual worlds in which virtual camera paths follow
those of the real world seed sequences to generate outputs we call
\emph{synthetic video clones}, which closely resemble the real-world data.
To build Virtual KITTI, we select five training videos from the original KITTI
MOT benchmark as ``real-world seeds'' to create our virtual worlds (\cf
Figure~\ref{fig:allclones}): 0001 (crowded urban area), 0002 (road in urban
area then busy intersection), 0006 (stationary camera at a busy intersection),
0018 (long road in the forest with challenging imaging conditions and shadows),
and 0020 (highway driving scene).

We decompose a scene into different visual components, with which off-the-shelf 
computer graphics engines (\eg game engines) and graphic assets (\eg geometric and
material models) can be scripted to reconstruct the scene.
We use the commercial computer graphics engine
Unity\footnote{\url{http://unity3d.com}} to create virtual worlds that closely
resemble the original ones in KITTI. This engine has a strong community that
has developed many ``assets'' publicly available on Unity's Asset Store.  These
assets include realistic 3D models and materials of objects.  This
allows for efficient crowd-sourcing of most of the manual labor in the initial
setup of our virtual worlds, making the creation of each virtual world
efficient (approximately one-person-day in our experiments).

The positions and orientations of the objects of interest in the 3D virtual
world are calculated based on their positions and orientations relative to the
camera and the position and orientation of the camera itself, both
available from acquired real-world data in the case of KITTI.
The main roads are also placed according to the camera position, with
minor manual adjustment in special cases (\eg the road changing width). 
To build the Virtual KITTI dataset, we manually place secondary roads and other
background objects such as trees and buildings in the virtual world, both
for simplicity and because of the lack of position data for them. Note that this
could be automated using Visual SLAM or semantic segmentation.
A directional light source together with ambient light simulates the sun. Its
direction and intensity are set manually by comparing the brightness and the
shadows in the virtual and real-world scenes, a simple process that only takes
a few minutes per world in our experiments.

\subsection{Changing conditions in synthetic videos}
\label{s:synthmods}

After the 3D virtual world is created, we can automatically generate not only
the clone synthetic video, but also videos with changed components.
This allows for the quantitative study of the impact of single factors (``ceteris
paribus analysis''), including rare events or difficult to observe conditions
that might occur in practice (``what-if analysis'').

The conditions that can be changed to generate new synthetic videos include
(but are not limited to): (i) the number, trajectories, or speeds of cars, (ii)
their sizes, colors, or models, (iv) the camera position, orientation, and
path, (v) the lighting and weather conditions.
All components can be randomized or modified ``on demand'' by changing
parameters in the scripts, or by manually adding, modifying, or removing
elements in the scene.

\begin{figure}
\center
\hspace*{-4mm}
\includegraphics[width=0.51\textwidth]{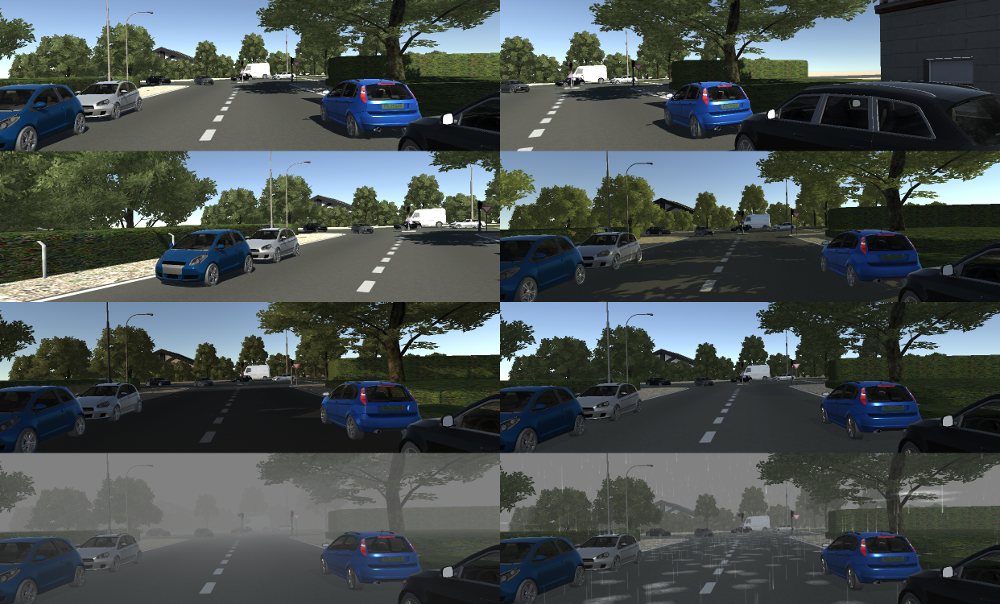}
\vspace*{-4mm}
\caption{\label{fig:weather} Simulated conditions. From top left to
bottom right: clone, camera rotated to the right by 15$^{\circ}$,
to the left by 15$^{\circ}$, ``morning'' and ``sunset'' times of day,
overcast weather, fog, and rain.} 
\vspace*{-2mm}
\end{figure}

To illustrate some of the vast possibilities, Virtual KITTI includes some
simple changes to the virtual world that translate in complex visual changes
that would otherwise require the costly process of re-acquiring and
re-annotating data in the real-world. 
First, we turned the camera to the right and then to the left, which lead to
some considerable change of appearances of the cars. Second, we changed
lighting conditions to simulate different time of the day: early morning and
before sunset. Third, we used special effects and a particle system together
with changed lighting conditions to simulate different weather: overcast, fog
and heavy rain.  See Figure~\ref{fig:weather} for an illustration.

\begin{figure*}
\center
\includegraphics[width=1.0\textwidth]{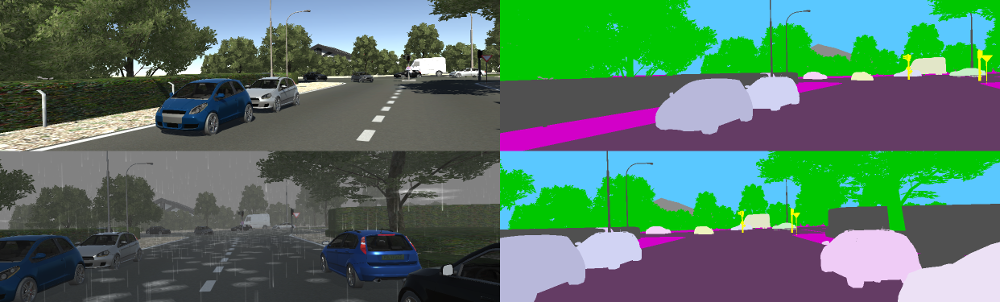}
\vspace*{-4mm}
\caption{\label{fig:pixel_gt} Rendered frame (left) and automatically generated
scene and instance-level segmentation ground-truth (right) for two modified
conditions: camera horizontally rotated to the left (top), rain (bottom).} 
\end{figure*}

\subsection{Generating ground-truth annotations}
\label{s:synthgt}

As stated above, ground-truth annotations are essential for computer vision
algorithms. In the KITTI dataset, the 2D bounding boxes used for evaluation
were obtained from human annotators by drawing rectangular boxes on the video
frames and manually labeling the truncation and occlusion states of objects.
This common practice is however costly, does not scale to large volumes of
videos and pixel-level ground-truth, and incorporates varying degrees of
subjectiveness and inconsistency.
For example, the bounding boxes are usually slightly larger than the cars and
the margins differ from one car to another and from one annotator to another.
The occlusion state (``fully visible'', ``partly occluded'', or ``largely
occluded'') is also subjective and the underlying criterion may differ
from case to case, yielding many important edge cases (occluded and truncated
cars) with inconsistent labels.

In contrast, our approach can automatically generate accurate and consistent
ground-truth annotations accompanying synthetic video outputs, and the
algorithm-based approach allows richer (\eg pixel-level) and more consistent
results than those from human annotators. 
We render each moment of the scene four times. 
First, we do the photo-realistic rendering of the clone scene by leveraging the
modern rendering engine of Unity.
Second, the depth map is rendered by using the information stored in the depth
buffer.
Third, the per-pixel category- and instance-level ground-truth is efficiently
and directly generated by using unlit shaders on the materials of the objects.
These modified shaders output a color which is not affected by the lighting and
shading conditions. A unique color ID is assigned for each object of interest
(\cf Figure~\ref{fig:pixel_gt}).
Fourth, we compute the dense optical flow between the previous and the
current frames by sending all Model, View, and Projection matrices for each
object to a vertex shader, and interpolate the flow of each pixel using a
fragment shader.
Note that these multiple renderings are an efficient strategy to generate
pixel-level ground truth, as it effectively leverages shaders offloading
parallel computations to GPUs (most of the computation time is used to swap
materials).
For Virtual KITTI, with a resolution of around $1242 \times 375$, the full
rendering and ground truth generation pipeline for segmentation, depth, and
optical flow runs at 5-8 FPS on a single desktop with commodity hardware.

We generate 2D multi-object tracking ground truth by (i) doing the perspective
projection of the 3D object bounding boxes from the world coordinates to the
camera plane (clipping to image boundaries in the case of truncated objects),
(ii) associating the bounding boxes with their corresponding object IDs to
differentiate object instances, and (iii) adding truncation and occlusion
meta-data as described below.
The truncation rate is approximated by dividing the volume of an object's 3D
bounding box by the volume of the 3D bounding box of the visible part (computed
by intersecting the original bounding box with the camera frustum planes).
We also estimate the 2D occupancy rate of an object by dividing the number of
ground-truth pixels in its segmentation mask by the area of the projected 2D
bounding box, which includes the occluder, as it results from the perspective
projection of the full 3D bounding box of the object.
In the special case of fog, we additionally compute the visibility of each
object from the fog formula used to generate the effect. 
%
To have comparable experimental protocols and reproducible ground truth
criteria across real and virtual KITTI, we remove manually
annotated ``DontCare'' areas from the original KITTI training ground truth
(\ie they may can count as false alarms), and ignore all cars smaller than
$25$ pixels or heavily truncated / occluded during evaluation (as described
in~\cite{Geiger2012}).
We set per sequence global thresholds on occupancy and truncation rates of
virtual objects to be as close as possible to original KITTI annotations.

\subsection{Assessing the usefulness of virtual worlds}
\label{s:transfer}

In addition to our data generation and annotation methods, a
key novel aspect of our approach consists in the assessment of the usefulness
of the generated virtual worlds for computer vision tasks.
It is a priori unclear whether and when using photo-realistic synthetic videos
is indeed a valid alternative to real-world data for computer vision
algorithms.
The \emph{transferability of conclusions} obtained on synthetic data is
likely to depend on many factors, including the tools used (especially
graphics and physics engines), the quality of implementation (\eg the
degree of photo-realism and details of environments and object designs or
animations), and the target video analysis tasks.
Although using synthetic training data is common practice in computer vision,
we are not aware of related works that systematically study \emph{the reverse},
\ie using real-world \emph{training} data, which can be noisy or weakly
labeled, and synthetic \emph{test} data, which must be accurately labeled and
where, therefore, synthetic data has obvious benefits.

To assess robustly whether the \emph{behavior} of a recognition algorithm is
similar in real and virtual worlds, we propose to compare its performance on
the initial ``seed'' real-world videos and their corresponding virtual world
clones.  We compare multiple metrics of interest (depending on the target
recognition task) obtained with fixed hyper-parameters that maximize
recognition performance on both the real and virtual videos, while
simultaneously minimizing the performance gap.
In the case of MOT, we use Bayesian hyper-parameter
optimization~\cite{bergstra2013making} to find fixed tracker
hyper-parameters for each pair of real and clone videos. We use as
objective function the sum of the multi-object tracking accuracies
(MOTA~\cite{Bernardin2008}) over original real-world videos and their
corresponding virtual clones, minus their absolute differences, normalized
by the mean absolute deviations of all other normalized CLEAR MOT
metrics~\cite{Bernardin2008}.

This allows us to quantitatively and objectively measure the impact of the
virtual world design, the degree of photo-realism, and the quality of other
rendering parameters on the algorithm performance metrics of interest. Note
that this simple technique is a direct benefit of our virtual world
generation scheme based on synthetically cloning a small set of real-world
sensor data.
Although the comparisons depend on the tasks of interest, it is also
possible to complement task-specific metrics with more general measures of
discrepancy and domain mismatch measures~\cite{Gretton2012}.

Finally, note that our protocol is complementary to the more standard approach
consisting of using synthetic training data and real-world test data.
Therefore, in our experiments with Virtual KITTI we investigate both methods
to assess the usefulness of virtual data, both for learning virtual models
applied in the real world and for evaluating real-world pre-trained models in
both virtual and real worlds.

\section{Experiments}
\label{s:experiments}

In this section, we first describe the MOT models used in our experiments.
We then report results regarding the differences between the original
real-world KITTI videos and our virtual KITTI clones.
We then report our experiments on learning in virtual worlds models applied on
real-world data.
Finally, we conclude with experiments to measure the impact of camera,
lighting, and weather on recognition performance of real-world pre-trained MOT
algorithms.

\subsection{Strong Deep Learning Baselines for MOT}
\label{s:motbaselines}

Thanks to the recent progress on object detection, association-based
tracking-by-detection in monocular video streams is particularly successful and
widely used for MOT~\cite{Breitenstein2011, Pirsiavash2011, Milan2014,
Geiger2014, Hall2014, Collins2014, Gaidon2015a, Wang2015e, Choi2015,Xiang2015}
(see~\cite{Luo2014} for a recent review). These methods consist in building
tracks by linking object detections through time.

In our experiments, the detector we use is the recent Fast-R-CNN object
detector from Girshick~\cite{Girshick2015} combined with the efficient Edge
Boxes proposals~\cite{Zitnick2014}.
In all experiments (except for the virtual training ones), we follow the
experimental protocol of~\cite{Girshick2015} to learn a powerful VGG16-based
Fast-RCNN car detector by fine-tuning successively from ImageNet, to Pascal VOC
2007 cars, to the KITTI object detection benchmark training
images\footnote{\url{http://www.cvlibs.net/datasets/kitti/eval_object.php}}.

To use this detector for association-based MOT, we consider two trackers.
The first is based on the principled network flow algorithm of~\cite{Zhang2008,
Pirsiavash2011}, which does not require video training data.
The maximum a posteriori (MAP) data association problem can indeed be elegantly
formalized as a special integer linear program (ILP) whose global optimum can
be found efficiently using max-flow min-cost network flow
algorithms~\cite{Zhang2008, Pirsiavash2011}.
In particular, the dynamic programming min-cost flow (DP-MCF) algorithm of
Pirsiavash~\etal~\cite{Pirsiavash2011} is well-founded and particularly
efficient. Although it obtains poor results on the KITTI MOT
benchmark~\cite{KITTIMOTres}, it can be vastly improved by (i) using a better
detector, (ii) replacing the binary pairwise costs in the network by using the
intersection-over-union, and (iii) allowing for multiple time-skip connections
in the network to better handle missed detections.
Our DP\_MCF\_RCNN tracker reaches $57\%$ MOTA on the KITTI MOT evaluation
server~\cite{KITTIMOTres}, improving by $+20\%$ \wrt the original
DP\_MCF~\cite{Pirsiavash2011}.
Note that this baseline tracker could be further improved, as shown recently by
Wang and Fowlkes~\cite{Wang2015e}. Their method indeed obtains $77\%$ MOTA with
a related algorithm thanks to better appearance and motion modeling coupled
with structured SVMs to learn hyper-parameters on training videos.

The second tracker we consider is the recent state-of-the-art Markov Decision
Process (MDP) method of Xiang~\etal~\cite{Xiang2015}. It relies on
reinforcement learning to learn a policy for data association from ground truth
training tracks. This method reaches $76\%$ MOTA on the KITTI MOT test set
using ConvNet-based detections.
In our experiments requiring a pre-trained tracker, we learned the MDP
parameters on the following 12 real-world KITTI training videos: 0000, 0003,
0004, 0005, 0007, 0008, 0009, 0010, 0011, 0012, 0014, 0015. (The remaining
videos are either the seed sequences used to create the virtual worlds, or
sequences containing no or very few cars.)

\def\mota{MOTA$\uparrow$}
\def\motp{MOTP$\uparrow$}
\def\mt{MT$\uparrow$}
\def\ml{ML$\uparrow$}
\def\frag{F$\downarrow$}
\def\ids{I$\downarrow$}
\def\fone{F1$\uparrow$}
\def\prec{P$\uparrow$}
\def\rec{R$\uparrow$}
\def\vr{v$\rightarrow$r}

\begin{figure*}
\centering
\vspace*{-4mm}
\hspace*{-6mm}
\resizebox{1.07\textwidth}{!}{%
\includegraphics[width=0.49\textwidth]{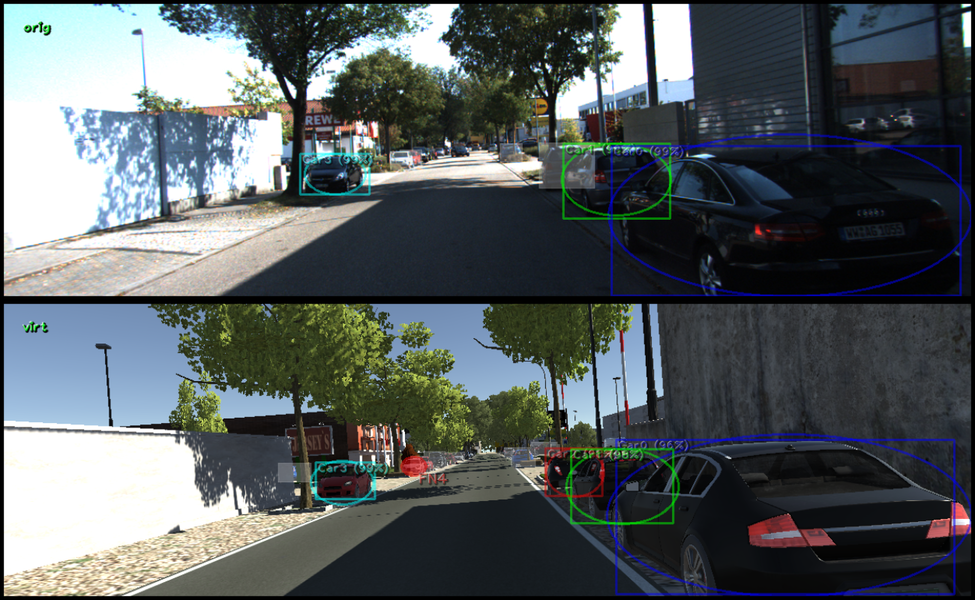} \
\includegraphics[width=0.49\textwidth]{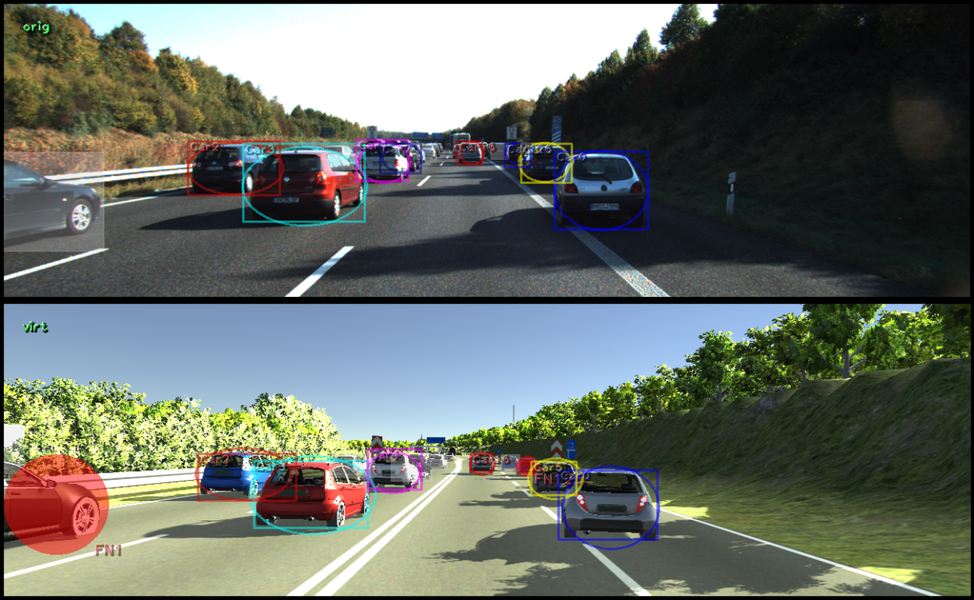}
}
\vspace*{-6mm}
\caption{\label{fig:trackclone} Predicted tracks on matching frames of two
original videos (top) and their synthetic clones (bottom) for both DP-MCF
(left) and MDP (right). Note the visual similarity of both the scenes and the
tracks. Most differences are on occluded, small, or truncated objects.} 
\end{figure*}

\begin{table*}
\hspace*{-7mm}
\centering
\resizebox{1.07\textwidth}{!}{%
\begin{tabular}{crrrrrrrr}
\rowcolor{gray!50}
DPMCF  &  \mota & \motp  &  \mt    &   \ml &\ids &\frag  & \prec  &  \rec   \\
\toprule                                                                      
0001    & 73.9\% & 86.8\% &  73.3\% &  4.0\%&  38 &   52  & 93.1\% & 85.5\% \\
v0001   & 73.1\% & 82.0\% &  58.2\% &  5.1\%&  30 &   47  & 98.4\% & 79.5\% \\
\midrule                                                                      
0002    & 72.5\% & 84.1\% &  54.5\% & 27.3\%&   4 &   20  & 99.6\% & 75.5\% \\
v0002   & 74.0\% & 78.7\% &  50.0\% & 20.0\%&   5 &   16  & 98.6\% & 77.9\% \\
\midrule                                                                      
0006    & 88.3\% & 85.6\% &  90.9\% &  0.0\%&   1 &   12  & 98.8\% & 90.6\% \\
v0006   & 88.3\% & 83.6\% & 100.0\% &  0.0\%&   3 &    7  & 94.6\% & 95.4\% \\
\midrule                                                                      
0018    & 93.0\% & 87.2\% &  82.4\% &  0.0\%&   1 &    7  & 95.2\% & 98.7\% \\
v0018   & 93.7\% & 73.0\% &  66.7\% &  0.0\%&   2 &   16  & 99.9\% & 94.4\% \\
\midrule                                                                      
0020    & 81.0\% & 84.8\% &  68.6\% &  4.7\%&  88 &  150  & 94.4\% & 90.0\% \\
v0020   & 82.0\% & 77.4\% &  44.8\% & 14.6\%&  67 &  142  & 99.3\% & 86.1\% \\
\midrule                                                                      
AVG     & 81.7\% & 85.7\% &  73.9\% &  7.2\%&  26 &   48  & 96.2\% & 88.1\% \\
v-AVG   & 82.2\% & 78.9\% &  63.9\% &  7.9\%&  21 &   45  & 98.2\% & 86.7\% \\
\bottomrule
\end{tabular}%
\quad
\begin{tabular}{crrrrrrrr}
\rowcolor{gray!50}
MDP   &  \mota & \motp  &  \mt   &   \ml  &\ids &\frag & \prec  &  \rec   \\ 
\toprule                                                                    
0001  & 81.8\% & 85.3\% & 78.7\% & 13.3\% &   5 &    6 &  91.1\% & 92.5\% \\
v0001 & 82.8\% & 81.9\% & 63.3\% & 13.9\% &   1 &   10 &  98.7\% & 85.8\% \\
\midrule                                                                    
0002  & 80.7\% & 82.2\% & 63.6\% & 27.3\% &   0 &    1 &  99.0\% & 82.5\% \\
v0002 & 81.1\% & 81.8\% & 60.0\% & 20.0\% &   0 &    2 &  98.4\% & 83.4\% \\
\midrule                                                                    
0006  & 91.3\% & 84.3\% & 72.7\% &  9.1\% &   0 &    3 &  99.7\% & 92.3\% \\
v0006 & 91.3\% & 84.4\% & 81.8\% &  9.1\% &   1 &    2 &  99.9\% & 92.0\% \\
\midrule                                                                    
0018  & 91.1\% & 87.0\% & 52.9\% & 35.3\% &   1 &    1 &  96.7\% & 95.2\% \\
v0018 & 90.9\% & 74.9\% & 44.4\% & 33.3\% &   0 &    0 &  99.1\% & 92.4\% \\
\midrule                                                                    
0020  & 84.4\% & 85.1\% & 58.1\% & 25.6\% &  14 &   24 &  96.7\% & 88.7\% \\
v0020 & 84.0\% & 79.4\% & 52.1\% & 34.4\% &   1 &    9 &  99.3\% & 85.6\% \\
\midrule                                                                    
AVG   & 85.9\% & 84.8\% & 65.2\% & 22.1\% &   4 &    7 &  96.7\% & 90.3\% \\
v-AVG & 86.0\% & 80.5\% & 60.3\% & 22.1\% &   0 &    4 &  99.1\% & 87.9\% \\
\bottomrule
\end{tabular}%
}
\vspace*{1mm}
\caption{DP-MCF (left) and MDP (right) MOT results on original real-world KITTI
train videos and virtual world video ``clones'' (prefixed by a ``v''). AVG
(resp. v-AVG) is the average over real (resp. virtual) sequences. We report the
CLEAR MOT metrics~\cite{Bernardin2008} -- including MOT Accuracy (MOTA), MOT
Precision (MOTP), ID Switches (I), and Fragmentation (F) -- complemented by the
Mostly Tracked (MT) and Mostly Lost (ML) ratios, as well as our detector's
precision (P) and recall (R).  }
\label{tab:resmot}
\vspace*{-5mm}
\end{table*}

\subsection{Transferability across the Real-to-Virtual Gap}
\label{s:mottransfer}

Table~\ref{tab:resmot} contains the multi-object tracking performance of
our DP-MCF and MDP trackers on the virtual KITTI clone videos and their
original KITTI counterparts following the protocol described in
Section~\ref{s:transfer}. See Figure~\ref{fig:trackclone} for some
tracking visualizations.

According to the MOTA metric which summarizes all aspects of MOT, \emph{the
real-to-virtual performance gap is minimal for all real sequences and their
respective virtual clones and for all trackers}, and $<0.5\%$ on average
for both trackers.
All other metrics show also a limited gap.
Consequently, the visual similarity of the sequences and the comparable
performance and behavior of the tracker across real-world videos and their
virtual worlds counterpart suggest that similar causes in the real and
virtual worlds are likely to cause similar effects in terms of recognition
performance. The amount of expected ``transferability of conclusions'' from
real to virtual and back can be quantified by the difference in the metrics
reported in table~\ref{tab:resmot}.

The most different metrics are the MOTP (average intersection-over-union of
correct tracks with the matching ground truth), and the fraction of Mostly
Tracked (MT) objects (fraction of ground truth objects tracked at least
$80\%$ of the time), which are both generally lower in the virtual world.
The main factor explaining this gap lies in the inaccurate and
inconsistent manual annotations of the frequent ``corner cases'' in the
real world (heavy truncation or occlusion, which in the original KITTI
benchmark is sometimes labeled as ``DontCare'', ignored, or
considered as true positives, depending on the annotator).
In contrast, our Virtual KITTI ground truth is not subjective, but
automatically determined by thresholding the aforementioned computed
occupancy and truncation rates.
This discrepancy is illustrated in Figure~\ref{fig:trackclone}, and
explains the small drop in recall for sequences 0001, 0018, and 0020 (which
contain many occluded and truncated objects).
Note, however, that the Fast-RCNN detector achieves similar F1 performance
between real and virtual worlds, so this drop in recall is generally
compensated by an increase in precision.

\subsection{Virtual Pre-Training}
\label{s:virtualpretraining}

As mentioned previously, our method to quantify the gap between real and
virtual worlds from the perspective of computer vision algorithms is
complementary to the more widely-used approach of leveraging synthetic data
to train models applied in real-world settings.
Therefore, we additionally conduct experiments to measure the usefulness of
Virtual KITTI to train MOT algorithms.

We evaluated three different scenarios: (i) training only on the 5 real
KITTI seed sequences (configuration '\textbf{r}'), (ii) training only on
the corresponding 5 virtual KITTI clones (configuration '\textbf{v}'), and
(iii) training first on the Virtual KITTI clones, then fine-tuning on the
real KITTI sequences, a special form of virtual data augmentation we call
\emph{virtual pre-training} (configuration '\textbf{\vr}').
We split the set of real KITTI sequences not used during training in two:
(i) a test set of 7 long diverse videos (4,5,7,8,9,11,15) to evaluate
performance, and (ii) a validation set of 5 short videos (0,3,10,12,14)
used for hyper-parameter tuning.
The Fast-RCNN detector was always pre-trained on ImageNet.
%
The MDP association model is trained from scratch using reinforcement
learning as described in~\cite{Xiang2015}.

\begin{table}
\vspace*{-3mm}
\hspace*{-4mm}
\centering
\resizebox{1.07\columnwidth}{!}{%
\begin{tabular}{lrrrrrrrr}
\rowcolor{gray!50}
         &  \mota & \motp  &  \mt  &   \ml &\ids &\frag & \prec  &  \rec  \\
\toprule
DP-MCF v & 64.3\% & 75.3\% & 35.9\% & 31.5\%&   0 &   15& 96.6\% & 71.0\% \\
DP-MCF r & 71.9\% & 79.2\% & 45.0\% & 24.4\%&   5 &   17& 98.0\% & 76.5\% \\
DP-MCF \vr  & 76.7\% & 80.9\% & 53.2\% & 12.3\%&   7 &   27& 98.3\% & 81.1\% \\
\midrule
MDP v    & 63.7\% & 75.5\% & 35.9\% & 36.9\% &   5 &   12 & 96.0\% & 70.6\% \\
MDP r    & 78.1\% & 79.2\% & 60.7\% & 22.0\% &   3 &    9 & 97.3\% & 82.5\% \\
MDP \vr  & 78.7\% & 80.0\% & 51.7\% & 19.4\% &   5 &   10 & 98.3\% & 82.6\% \\
\bottomrule
\end{tabular}%
}
\vspace*{1mm}
\caption{DP-MCF and MDP MOT results on seven held-out original real-world KITTI
train videos (4,5,7,8,9,11,15) by learning the models on \textbf{(r)} the five
real seed KITTI videos (1,2,6,18,20), \textbf{(v)} the corresponding five
Virtual KITTI clones, and \textbf{(\vr)} by successively training on the
virtual clones then the real sequences (virtual pre-training).
See Table~\ref{tab:resmot} for details about the metrics.
}
\label{tab:restrain}
\vspace*{-4mm}
\end{table}

Table~\ref{tab:restrain} reports the average MOT metrics on the
aforementioned real test sequences for all trackers trained with all
configurations.
Although training only on virtual data is not enough, we can see that the
best results are obtained with configuration \vr.  Therefore, \emph{virtual
pre-training improves performance}, which further confirms the usefulness
of virtual worlds for high-level computer vision tasks.
The improvement is particularly significant for the DP-MCF tracker, less
for the MDP tracker. MDP can indeed better handle missed detections and
works in the high-precision regime of the detector (the best minimum
detector score threshold found on the validation set is around $95\%$),
which is not strongly improved by the virtual pre-training.
On the other hand, DP-MCF is more robust to false positives but requires
more recall (validation score threshold around $60\%$), which is
significantly improved by virtual pre-training.
In all cases, we found that validating an early stopping criterion (maximum
number of SGD iterations) of the second fine-tuning stage of the \vr
\ configuration is critical to avoid overfitting on the small real training
set after pre-training on the virtual one.

\subsection{Impact of Weather and Imaging Conditions}
\label{s:motvarexps}

\begin{table}
\vspace*{-3mm}
\hspace*{-3mm}
\centering
\resizebox{1.07\columnwidth}{!}{%
\begin{tabular}{crrrrrrrr}
\rowcolor{gray!50}
DP-MCF   &  \mota & \motp  &  \mt    &   \ml &\ids &\frag  & \prec  &  \rec    \\
\toprule                                                                         
clone    & 82.2\% & 78.9\% &  63.9\% &  7.9\%&  21 &   45  & 98.2\% & 86.7\%   \\
\midrule
+15deg   &  -2.9\% & -0.8\% & -10.6\% &  6.3\% & -18 &  -31 &  0.5\% &  -3.9\% \\
-15deg   &  -8.1\% & -0.6\% &  -6.9\% & -1.9\% &  -8 &   -9 & -3.4\% &  -3.7\% \\
morning  &  -2.8\% & -0.3\% &  -6.0\% &  1.7\% &  -2 &   -3 &  1.0\% &  -3.9\% \\
sunset   &  -6.8\% & -0.0\% & -13.7\% &  3.6\% &  -2 &    0 & -0.6\% &  -6.1\% \\
overcast &  -2.0\% & -1.3\% & -12.3\% &  0.8\% &  -3 &   -5 &  0.5\% &  -2.7\% \\
fog      & -45.2\% &  4.0\% & -55.3\% & 33.3\% & -17 &  -29 &  1.1\% & -43.3\% \\
rain     &  -7.8\% & -0.4\% & -18.8\% &  3.3\% &  -9 &   -6 &  1.2\% &  -8.6\% \\
\bottomrule
\end{tabular}%
}%
\\
\vspace*{5mm}
\hspace*{-3mm}
\resizebox{1.07\columnwidth}{!}{%
\begin{tabular}{crrrrrrrr}
\rowcolor{gray!50}
MDP   &  \mota & \motp  &  \mt   &   \ml  &\ids &\frag & \prec  &  \rec        \\
\toprule
clone & 86.0\% & 80.5\% & 60.3\% & 22.1\% &   0 &    4 &  99.1\% & 87.9\%      \\
\midrule
+15deg   &  -5.9\% & -0.3\% &  -7.4\% &  6.2\% &   0 &    0 &  0.1\% &  -5.4\% \\
-15deg   &  -4.5\% & -0.5\% &  -4.8\% &  5.7\% &   0 &    3 & -0.5\% &  -4.0\% \\
morning  &  -5.1\% & -0.4\% &  -6.1\% &  3.1\% &   1 &    1 &  0.1\% &  -4.9\% \\
sunset   &  -6.3\% & -0.5\% &  -6.4\% &  4.3\% &   0 &    2 & -0.3\% &  -5.5\% \\
overcast &  -4.0\% & -1.0\% &  -7.2\% &  4.6\% &   0 &    0 & -0.2\% &  -3.6\% \\
fog      & -57.4\% &  1.2\% & -57.4\% & 40.7\% &   0 &   -2 & -0.0\% & -53.9\% \\
rain     & -12.0\% & -0.6\% & -15.3\% &  5.7\% &   1 &    3 & -0.2\% & -10.9\% \\
\bottomrule
\end{tabular}%
}
\vspace*{3mm}
\caption{Impact of variations on MOT performance in virtual KITTI for the
DP-MCF (top) and MDP (bottom) trackers. We report the average performance
on the virtual clones and the difference caused by the modified conditions
in order to measure the impact of several phenomena, all other things being
equal.  ``+15deg'' (resp. ``-15deg'') corresponds to a camera rotation of
$15$ degrees to the right (resp. left).  ``morning'' corresponds to typical
lighting conditions after dawn on a sunny day.  ``sunset'' corresponds to
slightly before night time.  ``overcast'' corresponds to lighting
conditions in overcast weather, which causes diffuse shadows and strong
ambient lighting. ``fog'' is implemented via a volumetric formula, and
``rain'' is a simple particle effect ignoring the refraction of water drops
on the camera.}
\label{tab:resmotvars}
\vspace*{-3mm}
\end{table}

Table~\ref{tab:resmotvars} contains the performance of our real-world
pre-trained trackers (Section~\ref{s:motbaselines}) in altered conditions
generated either by modifying the camera position, or by using special
effects to simulate different lighting and weather conditions.
As the trackers are trained on consistent ideal sunny conditions, all
modifications negatively affect all metrics and all trackers.  In
particular, bad weather (\eg fog) causes the strongest degradation of
performance. This is expected, but difficult to quantify in practice
without re-acquiring data in different conditions. This also suggests that
the empirical generalization performance estimated on the limited set of
KITTI test videos is an optimistic upper bound at best.
Note that the MDP tracker is suffering from stronger overfitting than
DP-MCF, as suggested by the bigger performance degradation under all
conditions.

\section{Conclusion}
\label{s:conclusion}

In this work we introduce a new fully annotated photo-realistic synthetic
video dataset called Virtual KITTI, built using modern computer graphics
technology and a novel real-to-virtual cloning method.
We provide quantitative experimental evidence suggesting that the gap
between real and virtual worlds is small from the perspective of high-level
computer vision algorithms, in particular deep learning models for
multi-object tracking.
We also show that these state-of-the-art models suffer from over-fitting,
which causes performance degradation in simulated modified conditions
(camera angle, lighting, weather).
Our approach is, to the best of our knowledge, the only one that enables to
scientifically measure the potential impact of these important phenomena on the
recognition performance of a statistical computer vision model.

In future works, we plan to expand Virtual KITTI by adding more worlds, and
by also including pedestrians, which are harder to animate.
We also plan to explore and evaluate domain adaptation methods and larger
scale virtual pre-training or data augmentation to build more robust models
for a variety of video understanding tasks, including multi-object tracking
and  scene understanding.

{\small
\bibliographystyle{unsrt}
\bibliography{references}
}

\end{document}